\def\year{2020}\relax
\documentclass[letterpaper]{article} % DO NOT CHANGE THIS
\usepackage{aaai20}  % DO NOT CHANGE THIS
\usepackage{times}  % DO NOT CHANGE THIS
\usepackage{helvet} % DO NOT CHANGE THIS
\usepackage{courier}  % DO NOT CHANGE THIS
\usepackage[hyphens]{url}  % DO NOT CHANGE THIS
\usepackage{graphicx} % DO NOT CHANGE THIS
\usepackage{ulem}
\usepackage{graphicx}  % DO NOT CHANGE THIS
\nocopyright
\usepackage{multirow}
\usepackage{booktabs} 
\usepackage[T1]{fontenc}

\title{A Unified Conversational Assistant Framework\\for Business Process Automation}
\author{Yara Rizk, Abhishek Bhandwalder, Scott Boag, Tathagata Chakraborti,\\ 
\Large \textbf{Vatche Isahagian, Yasaman Khazaeni, Falk Pollock, Merve Unuvar} \\ % All authors must be in the same font size and format. Use \Large and \textbf to achieve this result when breaking a line
\textsuperscript{}IBM Research AI\\ %If you have multiple authors and multiple affiliations
75 Binney Street\\
Cambridge, Massachusetts 02142\\
yara.rizk@ibm.com, abhi.b@ibm.com, scott\_boag@us.ibm.com, tchakra2@ibm.com,\\
vatchei@ibm.com, yasaman.khazaeni@us.ibm.com, falk.pollock@ibm.com, munuvar@us.ibm.com % email address must be in roman text type, not monospace or sans serif
}

\begin{document}

\maketitle

\begin{abstract}
Business process automation is a booming multi-billion-dollar industry that promises to remove menial tasks from workers' plates -- through the introduction of autonomous agents -- and free up their time and brain power for more creative and engaging tasks. However, an essential component to the successful deployment of such autonomous agents is the ability of business users to monitor their performance and customize their execution. A simple and user-friendly interface with a low learning curve is necessary to increase the adoption of such agents in banking, insurance, retail and other domains. As a result, proactive chatbots will play a crucial role in the business automation space. Not only can they respond to users' queries and perform actions on their behalf but also initiate communication with the users to inform them of the system's behavior. This will provide business users a natural language interface to interact with, monitor and control autonomous agents. In this work, we present a multi-agent orchestration framework to develop such proactive chatbots by discussing the types of skills that can be composed into agents and how to orchestrate these agents. Two use cases on a travel preapproval business process and a loan application business process are adopted to qualitatively analyze the proposed framework based on four criteria: performance, coding overhead, scalability, and agent overlap. 
\end{abstract}

% Discussion and comments Nov 1 2019 at 10.30-11:00
% mention skills can be RPAs... if they are RPA 2.0, we want to monitor them and invoke them
% how many dashboards can one look at all at once... our approach gives unified interface to link to different ones.
% learning components for 3S --> no need for uniform skill confidence, or not even provide confidence --> justify pipeline of orchestrator based on diversity of skill developers
% events trigger agents and o-base not just NL/convo

\section{Introduction}
Automation made its way into manufacturing plants in the early 1900s, however comparatively, its integration into the service industry has been slow. The service industry first took advantage of automation through customer care chatbots which routed incoming customer calls to the appropriate human operators. Gradually, these robotic answering machines improved and began to replace services provided by some human operators by walking customers through simple steps to resolve problems and escalating to humans only when necessary. As enabling technologies improved, so have these customer care chatbots which can now carry out more elaborate conversations with the customers.

Beyond chatbots, little automation was incorporated in the services industry despite the plethora of automation opportunities. The emergence of Robotic Process Automation (RPA) led to the next breakthrough. RPA developed robotic software agents capable of automating simple tasks within a business process such as form filling and data mining. RPAs reduced the cost of deploying autonomous agents but covered a small portion of automatable tasks, namely simple repetitive tasks. This led to the emergence of Business Process Automation (BPA) which supersedes RPA's scope. In addition to task automation, BPA looked to automate decision making, data management, content digitization, and workflow improvement. BPA relies on various artificial intelligence technologies including optical character recognition (OCR) systems and decision making models to automate business process tasks. As state-of-the-art algorithms become more accurate and reliable - deep learning OCR has achieved super-human accuracy levels - automation in service industry will become more cost-effective. 

Nevertheless, wide-spread deployment of autonomous systems has still lagged behind expectations. Trustworthiness, coding overhead, scalability, algorithm life cycle, technical expertise and user experience have been a few challenges that have stood in the way of more successful integration into business process solutions. In this work, we claim that a proactive multi-agent conversational framework can be that one stop shop that resolves many of these challenges. However, multiple research questions at the intersection of automation and business processes must be address first. A conversational interface reduces the necessary technical expertise to interact, monitor and customize the framework, but many algorithms do not have a conversational interface. How can we effectively integrate such algorithms into a conversational framework with minimal coding overhead? How can such independently developed (and potentially incompatible) functions coexist and cooperate within the same ecosystem? How could the user access all of the automation options in the framework with minimal interface and protocol hoping? The goal is not to create a natural language interaction for every part of the process but a single interactive interface. If it is accessing different documents, running OCR on a new application, checking our database or setting up alerts on specific events in our system, one would prefer to interact with a single interface which is conveniently conversational in this scenario.

We believe an orchestrator and agent composition approach with a specific pipeline would provide solutions to these questions. We present a multi-agent orchestration framework that consists of heterogeneous agents such as dialog, informational retrieval, task execution, and alerting agents and an orchestration layer that consists of three components (scorer, selector and sequencer) that would control the execution of the agents in the framework. As a proof of concept, we create two instances of this assistant for a travel preapproval business process and a loan application business process -- deployed in a sandbox environment -- and discuss how this framework can address these challenges.  

\section{Related Work}
\subsection{Business Process Automation}
Task automation has driven the digital transformation in enterprises with the emergence of RPA as a light-weight approach for automating repetitive tasks. RPA is a paradigm that seeks to automate the ``mouse-click" in user interfaces to relieve human resources of repetitive tasks. RPA adopts an outside-in approach that requires minimal changes to legacy software \cite{van2018robotic}; they have been applied to various sectors including accounting \cite{moffitt2018robotic}, auditing \cite{fernandez2018impacts}, human resources \cite{papageorgiou2018transforming}, banking \cite{stople2017lightweight}, public administration \cite{houy2019robotic} and energy sectors \cite{lacity2015robotic}. 
%RPA for recruiting \cite{nawaz2019robotic}

Multiple approaches have been adopted in the development of RPAs. \cite{gao2019automated} proposed an RPA solution for document flow automation in a debt collector business process. They used deep learning OCR and classification to control the path of the document through the business process. User behavior was captured to train a self-learning agent capable of identifying relationships between tasks and deploying RPAs based on the learned rules \cite{wroblewska2018robotic}. This approach, known as form-to-rule, relied on first-order logic to deduce the rules. Another branch of literature focused on identifying RPA-eligible tasks automatically. Leopold et al. relied on natural language processing of business process descriptions \cite{leopold2018identifying}. They adopted supervised machine learning to classify instances into one of three classes.  

As the scope of RPA increases to include more complex tasks, RPA agents incorporate other technological advancements such as OCR for document digitization, and data mining for content collection and management to successfully automate complex tasks. These advancements have achieved the best performance when adopting deep learning models \cite{schmidhuber2015deep}. However, these models are computationally expensive to train and have seen slow adoption despite their super-human accuracy due to their lack of transparency and explainability. 

Beyond RPA, BPA also seeks for automated decision making within business processes. The field of automated planning has been considered a prime candidate to enable more advanced levels of automation in business processes \cite{marrella2017automated}. Planning algorithms can be used to automatically generate business process models, adapt the business process' behavior based on unforeseen circumstances, and autonomously perform conformance checking. %Since planning models are generally more human understandable, they can be a good fit for BPA.  

End-to-end BPA has yet to find success in real-world deployments despite the ubiquity of computing devices and the advancements in software technology. Trust and reliability have been two of the main challenges preventing wide-spread adoption of automation approaches. With business stakeholders lacking the technical skills to monitor autonomous agents and intervene when they do not operate as expected, they have been reluctant to adopt this technology. Providing a natural language interface to allow such users to interact with autonomous agents may be the key to increase the adoption of automation in service enterprises.  

\subsection{Conversational Agents} 
Natural language is the main communication modality in business enterprises. From conversations to written documents, forms and emails, understanding and generating natural language can significantly facilitate the integration of various autonomous agents into domains where users do not speak the language of computers. Conversational agents have been developed for a wide range of applications using a plethora of machine learning techniques that have resulted in agents with increasing degrees of sophistication. In this section, we will focus on chatbots developed for enterprise applications and specifically those used in conjunction with RPA. Two approaches to chatbot development have been mainly adopted in the literature: 1) building domain specific chatbots by relying on natural language understanding and generative machine learning models, and 2) training deep learning models on data samples \cite{galitsky2019developing}. While the former requires more overhead to develop, it has been the more widely adopted approach for enterprise chatbots due to its predictable and controllable behavior. 

In enterprises, chatbots were first adopted for customer support which involved answering customers' questions about the services, products, location etc. of businesses \cite{galitsky2019developing}. Beyond question answering bots, a shopping bot was developed to increase the sales of shops by answering questions from customers and even providing custom prices for products \cite{heo2018chatbot}. A delivery bot was also developed for food delivery services that reduced the effort required from customers to order pizza \cite{heo2018chatbot}. While only automating 30\% of the processes with these bots, the bot developers have recently begun investigating the incorporation of RPA to increase the percentage of automation. Matthies et al. developed a chatbot for agile software development teams \cite{matthies2019additional} which analyzed commits in version control systems to provide insights into the teams' performance. Kalia et al. proposed an approach to generate a conversational agent based on dialog trees given a business process flow \cite{kalia2017quark}. Even though they provide a systematic way to develop chatbot that were sound, significant overhead and domain knowledge were necessary. 

Furthermore, these approaches have focused on designing a single conversational agent capable of answering frequently asked questions and performing tasks in a process such as determining the price of products in the shopping bot. Expanding the scope of their operation would require significant coding overhead by developers who understand the existing system. Also, reusing existing agents may not be possible due to compatibility reasons. In this work, we propose a conversation agent framework based on the multi-agent system paradigm as a potential solution to these issues. Instead of designing a single powerful agent, multiple specialized agents with narrower scopes can be independently designed and then integrated into a multi-agent system to act as a single conversational agent from the user's perspective. 

\section{Methodology} 

\subsection{Overall Framework}
\begin{figure}
    \centering
    \includegraphics[width=0.8\linewidth]{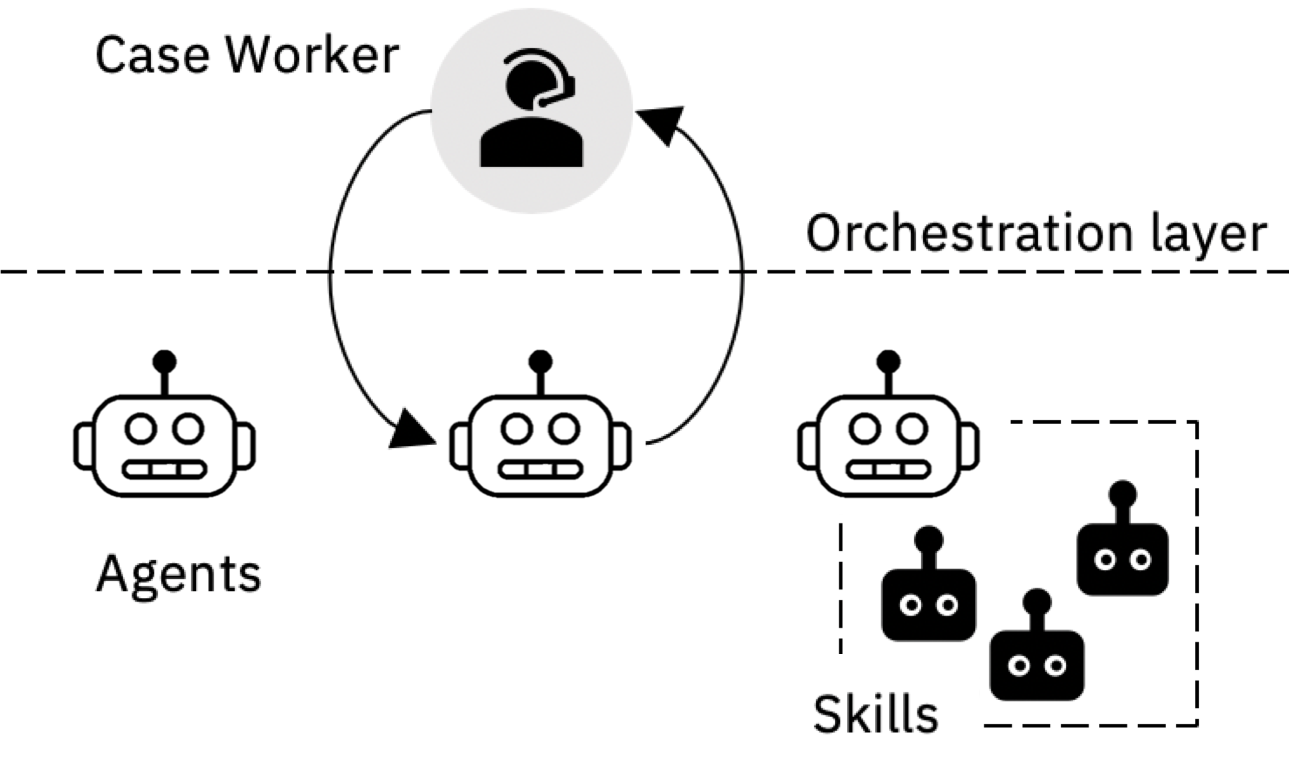}
    \caption{Multi-agent Orchestration Overview}
    \label{fig:MAOO}
\end{figure}

The framework for the proactive conversational assistant for BPA consists of three main components: skills, agents, and an orchestrator, as shown in Fig. \ref{fig:MAOO}. Skills perform well defined tasks; they require a set of inputs to produce a set of outputs. Agents are composed of skills that fall into three main categories: understand skills, act skills, and respond skills. An execution pattern for the skills must be provided; it can be as simple as a static sequential script or as complex as a full execution graph. The orchestrator coordinates between agents and it determines which agent or agents must execute to successfully respond to a user. When a user enters an utterance, this natural language sentence is forwarded to all the agents in the assistant by the orchestrator after potentially preprocessing the input. The orchestrator requests a preview of each agent's response and their confidence score. This score depicts how confident an agent is in its ability to respond to the user. Depending on the adopted orchestration pattern, the orchestrator selects one or more agents to execute and return their responses to the user. 

\subsection{Skills}
Skills are atomic functions, the building blocks of agents. They are categorized according to their role within an agent and their effect on the state of the world. Skills can have one of three roles: understand, act and respond. Furthermore, skills can either be world changing or non-world changing.

\subsubsection{Understand skills}
Serving as the entry point to an agent, understand skills identify the user's intent. In the conversational setting, understand skills are generally intent and entity recognition skills but can also be as simple as a key-phrase detection. They may take a user's natural language utterance and classify it into one of many predefined intents that the agent can handle or check if a specific phrase designated to trigger the agent is there. They may also annotate any entities they recognize within the input. In a non-dialog setting, understand skills are event triggered skills that watch for specific events and state changes in the system. These events could be a user logging in or a change in database. 
% Does slot filling fall into the entry skill? more like act skill cuz it executes after orchestrator selects the agent... a bit of overlap
% YR comment: more generally, some agents will not have an intent recognition entry skill; what could that entry skill be?

\subsubsection{Act skills}
Act skills process the user input to produce an output that can be used to generate a response. We distinguish two main types of act skills: world changing and non-world changing. World changing skills perform actions that modify the state of the world or have side-effects outside the confines of the orchestrator world. Examples of such skills include credit score checking skills, email sending skills, etc. Non-world changing skills do not have any side effects. Examples include email reading skills, weather checking skills, etc. 
Act skills in the business process world can perform various tasks or activities. One task is gathering information by querying databases or making API calls to information sources. Decision points are a set of activities that move the business process forward. Another task that moves the business process forward is submitting applications or forwarding gathered information to the next activity. 

\subsubsection{Respond skills}
Respond skills create the response that will be returned to the user by the conversational assistant. The skills take the output of act skills and process them to form natural language responses to the user; these could be full flesh natural language generation to template responses given different outputs of the other skills. Other modalities of responses can also be considered such as visualization of the output in the form of plots or images. 

\subsection{Agents}
Individually, the skills may not be able to perform complex tasks required by the activities within a business process. Instead of increasing the scope of individual skills by implementing more complex functionality within them, composing them into agents capable of completing these activities is more effective. This modular approach simplifies debugging of skills and agents and increases the reusability of skills by making them as domain agnostic as possible. 

We define a contract for the agents in order to facilitate their orchestration within a single assistant. The agents expect to receive an input message or utterance and the current context which represents the state of the system. The agent is capable of producing a preview response that does not cause any world-changing actions to be invoked, an execute response that results from fully executing its pipeline and may cause the world to change, an updated context once the agent has executed, and a confidence in its response.  

Multiple approaches can be adopted to compose skills into agents. Creating the execution pipeline can be done statically where a software developer writes a script that explicitly defines the order in which skills must be executed and directs the flow of data by routing inputs and outputs. In goal-oriented non-deterministic planning based composition \cite{muise2019planning,botea2019generating}, software developers declaratively define the inputs and outputs of skills and the end goal of the agent. Then, a planning engine composes an execution tree using the subset of available skills that will lead to the goal. While this approach allows independent developers to include their skills in a catalog, composing skills into agents in an effective way requires a minimum level of compatibility between the skills, including the adoption of a unified vocabulary for input/output variables.  

Agents are composed of an understand skill, a set of act skills, and a respond skill. Depending on the functionality of the agent, none, one or multiple act skills may be included in the composition. %YR comment: a figure here could be useful; maybe add it to agent anatomy fig
Next, we discuss four main types of agents based on their functionality. While many more agents can be created, we consider the ones that would be most relevant and commonly used in BPA. 

\subsubsection{Dialog Agents}
Users may just need to inquire about certain topics and do not require any action to be taken by the assistant. In such scenarios, a dialog agent, composed of an understand skill for intent and entity recognition and a respond skill to answer the user's query. Examples of such agents include chit-chat agents, FAQ or ``help" agents, etc. They can be implemented in dialog tree type services. These agents do not change the world; their preview mode and execute mode are the same. 

\subsubsection{Information Retrieval Agents}
Unlike dialog agents, information retrieval agents must connect to an external service (such as making an API call or querying a database) in order to produce their answer to a user's query as opposed to having the answering within the agent implementation. As a result, such agent compositions consist of three types of skills: understand skill that determine the intent of the user and annotate any entities that may be useful, one or more act skills that queries the appropriate service to produce its answer, and a respond skill that appropriately constructs the response returned to the user. In general, most agents within this category do not contain skills that change the world. However, their preview and execute modes may vary for computational/latency reasons. In some situations, querying does change the world; for example, retrieving the credit score of a loan applicant is considered as world changing as it will have a negative impact on the applicant if queried a large number of times. In other cases, simply incurring a cost, for example monetary or in the form or using up API calls that are limited, to query a database can be modeled as world changing to minimize the cost incurred by the system. 

\subsubsection{Task Execution Agents}
Beyond querying for information, the business user may want the assistant to perform tasks within the business process. Task execution agents are capable of moving the business process forward by submitting application, making decisions at decision points and such. These agents will change the world and therefore may require a preview and execute mode based on the type of orchestrator adopted. These agents are composed of an understand skill, a set of act skills and a respond skill. 

\subsubsection{Alerting Agents} %TODO: (YR) revisit this paragraph... doesn't flow well
Setting up alerts to receive notifications triggered by the occurrence of specific events allows business users to monitor the business process in an asynchronous way. Like the other agents, these agents are also composed of an understand skill, a set of act skills and a respond skill. However, their understand skill does not necessarily have to be a natural language understanding skill. However, setting up custom alerts would require an intent and entity recognition understand skill. These agents consist of a watcher skill on the system that may trigger the alert, a notification generation skill and a communication channel skill that sends the notification over the appropriate communication channel (e.g. email, SMS, slack...). A natural language understanding skill can be added to allow the user to customize the trigger and settings through dialog. If that is the case, a Natural Language to Query (NLQ) skill would be necessary to convert natural language trigger definitions to structured, well-defined queries on the database that may trigger the alert and is watched by the daemon. If an NLQ skill is not available, hard-coded conversions must be present. However, this approach restricts the number of alerts that can be created. NLQ provides a broader range of triggers that can be set. It also allows the creation of domain agnostic alert agents which would be more portable and require minimal coding overhead to deploy in new domains, a desirable feature in the BPA field. In this paper, we distinguish between alerts and notifications. The former is the occurrence of an event that causes the conditions of the trigger to be satisfied. The latter is the actual message sent to the user when an alert happens. 

\subsection{Orchestration}
Combining these diverse agents into a single assistant requires a domain agnostic orchestration layer capable of determining which subset of agents are best suited to respond to the user's utterance. Various orchestration frameworks with diverse characteristics can be adopted to achieve this goal. Stateless orchestrators do not keep track of the state in a centralized location. Instead the context variables (state) are passed between the orchestrator and agents. Stateful orchestrators on the other hand, maintain the state of the system in a centralized location. The computational cost of data transfer in the former approach increases with the number of context variables maintained in the state. As a result, system designers tend to minimize the number of variables in their state resulting a decrease in the amount of information agents have access to. Stateful orchestrators, however, require a central device to maintain the state which could be costly from a deployment and scalability perspective. 

Apriori orchestrators decide which agents should execute by considering the user input and what they know about agent capabilities. Posterior orchestrators pass the inputs to the agents requesting their responses, then decide which agents to select. In this case, preview and execute modes of agents must be available for those agents that perform world changing actions. Orchestrators also expect agents to provide a confidence in their responses which are used as a feature in their decision making algorithm. 

Unlike dialog managers, orchestrators do not require knowledge of the dialog to select agents. Dialog managers are assumed to be part of those agents that need it. For example, dialog agents would have managers that keep track of which node of a tree the conversation is in or perform dialog disambiguation and co-reference resolution. 

%\subsection{3S Orchestrator} % YR comment: still not sure if this should be here or in use case section
The 3S orchestrator is an example of a posterior orchestrator that relies on the agent's confidence, among other metrics, to determine which agents should respond. Its pipeline consists of three main steps: scoring, selecting, and sequencing agents \cite{yurochkin2019online,upadhyay2019bandit}. %In its current implementation, it is a stateless orchestrator. 

\begin{figure}
    \centering
    \includegraphics[width=0.8\linewidth]{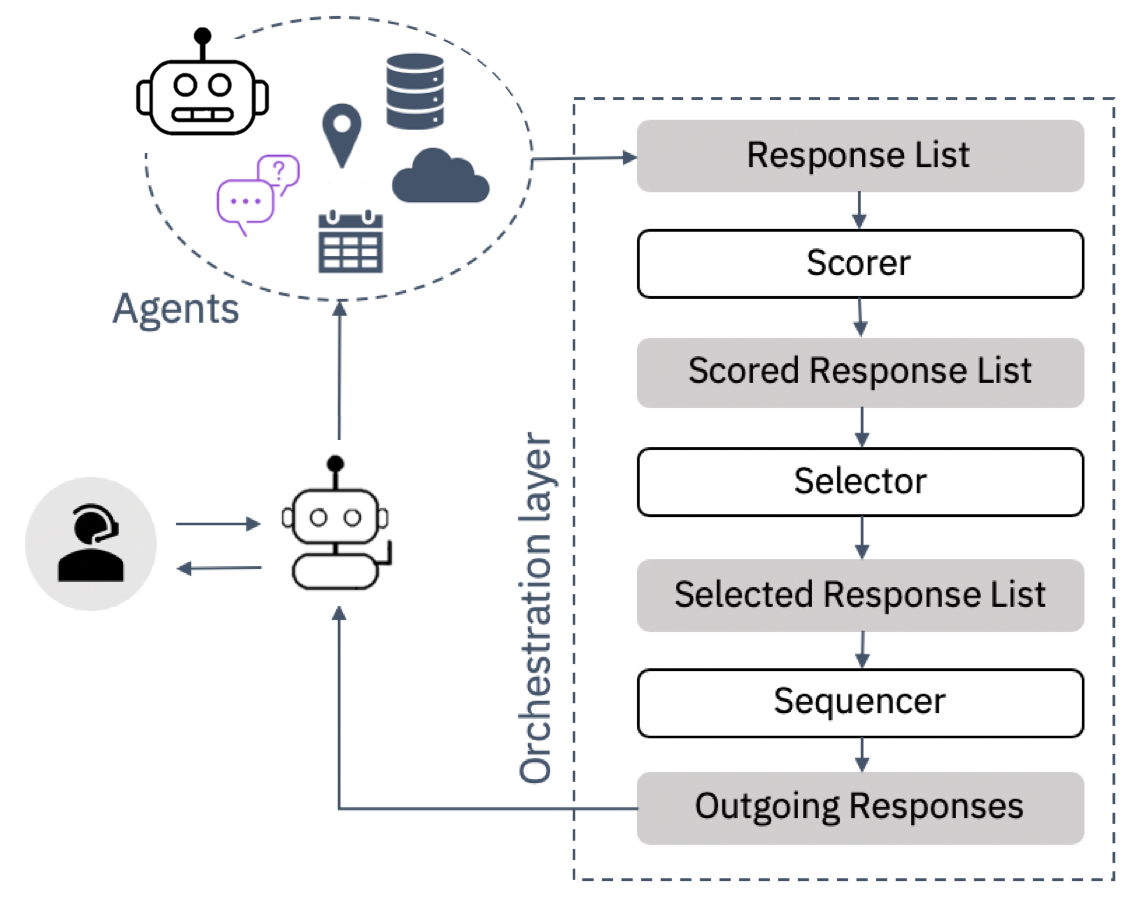}
    \caption{3S Orchestration Pipeline} % N.B.: After selector, agents must send execution message not preview message. TODO: update figure accordingly; remove example agents (just use abstract figs)?
    \label{fig:orchPipeline}
\end{figure}

% \begin{figure}
%     \centering
%     \includegraphics[width=\linewidth]{orch_workflow.png}
%     \caption{Framework Workflow} % N.B.: After selector, agents must send execution message not preview message. TODO: update figure accordingly; remove example agents (just use abstract figs)?
%     \label{fig:orchWorkflowSample}
% \end{figure}

\subsubsection{Scorer}
When the orchestrator receives the user's utterance, it forwards this input to all agents and requests a preview of their response. Once the agents return their preview responses and their confidence score, the scorer performs additional computations to obtain the final scores per agent. An identity function scorer simply forwards the scores as is to the selector. A scorer may normalize scores, especially since they could have been produced by very diverse agents that compute their scores differently. A more sophisticated scorer could enhance the agents' confidences with other measures such as conversation stickiness, dialog depth (within a dialog tree) and agent engagement. A Bayesian model that adapts to user feedback could also be adopted. 

\subsubsection{Selector}
Once the scorer computes the final scores per agent, it forwards them to the selector that must select one or more agents that will respond to the user. It can be as simple as a \textit{Top 1 (or Top K) score} selector that picks the agent (or K agents) with the highest score. More complex selection criteria could be adopted such as heuristic rules, learned selection models from labeled user data, online reinforcement learning models trained on user feedback, or a combination of multiple approaches. Once the selector determines the subset of agents that must execute, the orchestrator requests their execution responses. 
Various machine learning algorithms can be adopted to improve the robustness of the selector as the noise in the score increases. In a supervised learning domain, labels can represent agents that are the output of classifiers and the input to the classifier would be a feature vector consisting of the agents' scores and/or user utterances. In a reinforcement learning domain, actions can represent agents and the environment produces a reward when the correct agent is selected. For both approaches, deep learning models can be trained if enough data is provided to further improve the selection model. 

\subsubsection{Sequencer}
After receiving the agents' execution responses, a sequencer determines the order in which these responses will be presented to the user. If a \textit{Top 1 score} selector was adopted in the previous stage, then no sequencing is necessary. The response is directly outputted to the user through the conversational interface. However, when a subset of agents has responded, a sequencer could adopt different approaches to order these responses: e.g. descending score, heuristic rules (e.g. greeting agent must always respond before other agents), supervised or reinforcement learning models trained on labeled data and user feedback.

\section{Use Case -- 1: Travel Preapproval Application}
We will now detail the implementation of a conversational assistant (in Slack) for two personas (employees, and managers/directors) in a simplified travel preapproval application process using the framework described so far. 

% \begin{figure}
%     \centering
%     \includegraphics[width=\linewidth]{Travelbot_agents.png}
%     \caption{\textit{Travelbot} Agents} % TODO: (YR) add "response" skills to agents that are missing it
%     \label{fig:travelbot}
% \end{figure}

\begin{figure}
    \centering
    \includegraphics[width=0.9\linewidth]{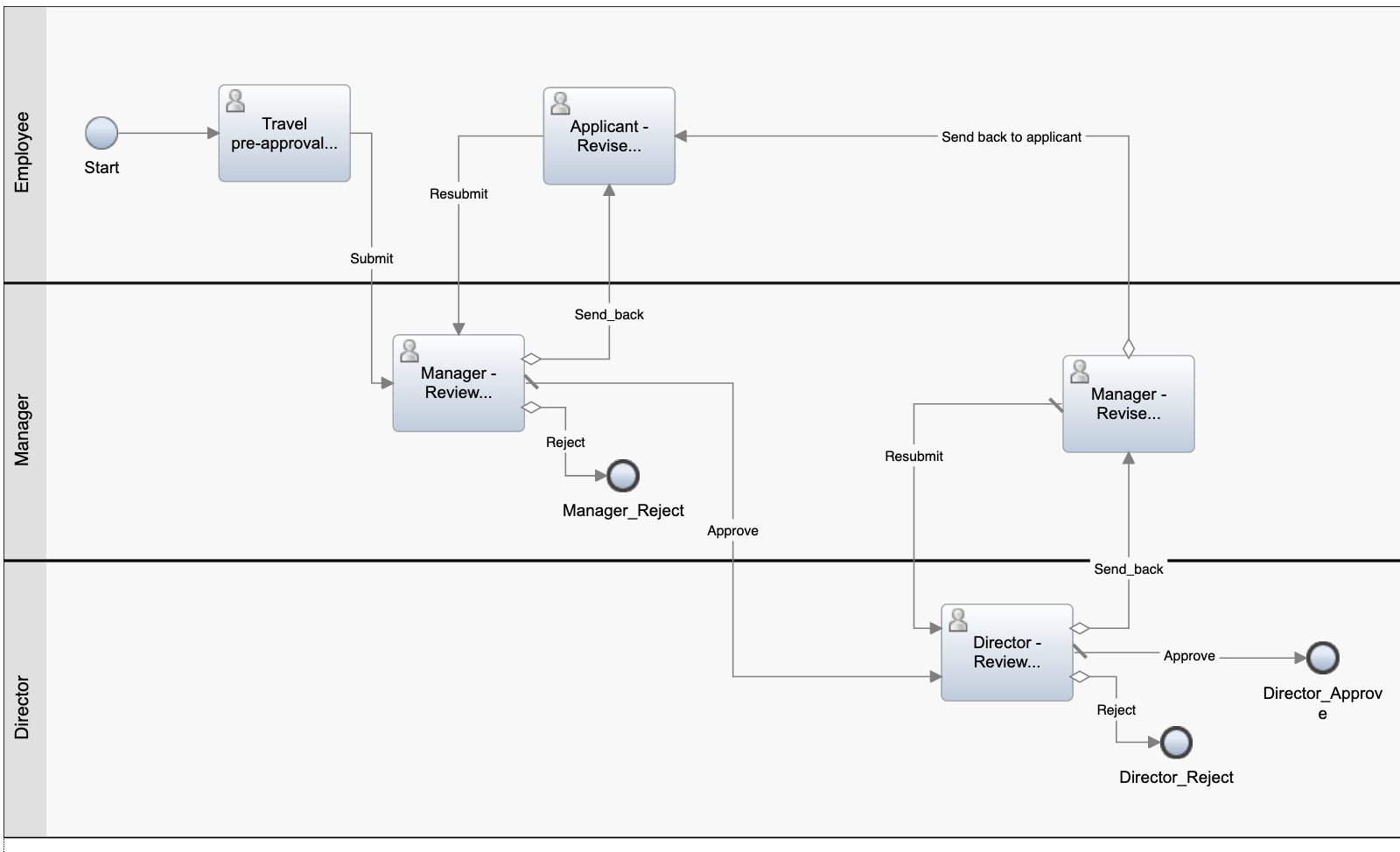}
    \caption{Travel Preapproval Process}
    \label{fig:bpm_travel}
\end{figure}

\begin{table}\small
    \centering
    \begin{tabular}{p{5.5cm} l}
         \textbf{Conversation} & \textbf{Responding agent} \\
         \toprule
         Manager: Hello & \multirow{2}{*}{Chit-Chat}\\
         \textit{Travelbot}: Hi there & {} \\
         \midrule
         Manager: Retrieve the number of accepted papers authored by John Smith & \multirow{2}{*}{Publication Query}\\
         \textit{Travelbot}: The number of accepted papers by John Smith is 7 & {} \\
        %  \midrule
        %  Manager: Plot this data as a bar graph & \multirow{2}{*}{Visualization}\\
        %  \textit{Travelbot}: Here is the plot (Fig. \ref{fig:viz_output}) & {} \\
         \midrule
         Manager: Approve John Smith's request & Business Process \\
         \textit{Travelbot}: John Smith's application has been approved & Task Execution \\
         \bottomrule
    \end{tabular}
    \caption{Conversation samples}
    \label{tab:travelbot_convo}
\end{table}

% \begin{figure}
%     \centering
%     \includegraphics[width=\linewidth]{Viz_output.png}
%     \caption{Visualization agent output}
%     \label{fig:viz_output}
% \end{figure}

\subsection{Travel Preapproval Business Process}
We adopt a simplified business process revolving around travel preapprovals generally used across many corporations. We consider the scenario of researchers/employees requesting travel funds to attend research conferences. The travel preapproval process, shown in Fig. \ref{fig:bpm_travel}, was implemented in a business process management software. 

\begin{figure}
    \centering
    \includegraphics[width=0.9\linewidth]{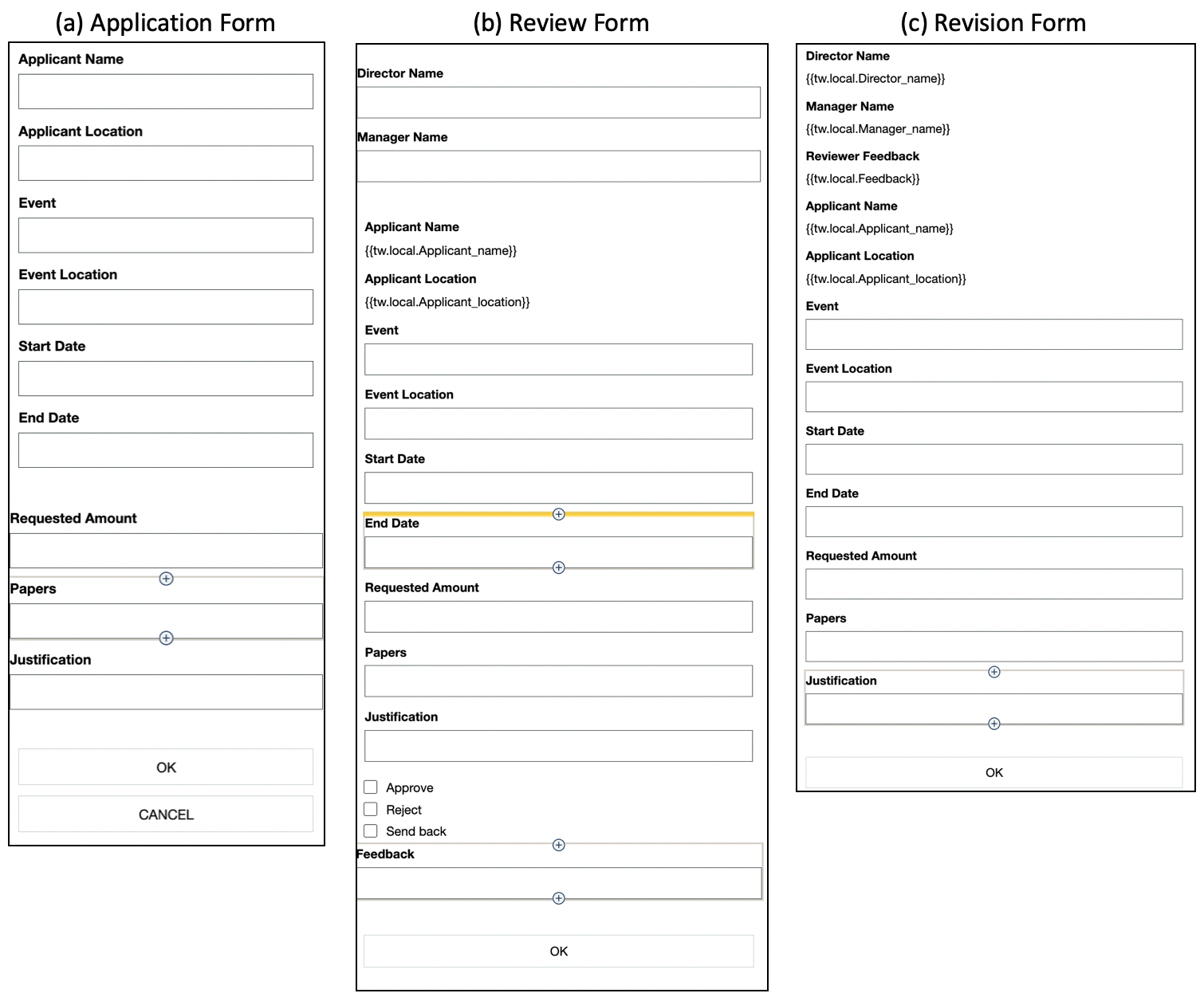}
    \caption{Employee Travel Preapproval User Interfaces: (a) Employee Application Form, (b) Manager/Director Review Form, (c) Employee/Manager/Director Revision Form}
    \label{fig:BPM_UIs}
\end{figure}

Employees submit a travel preapproval request to attend a conference by filling in specific information into a form, shown in Fig. \ref{fig:BPM_UIs}(a). Once submitted, the application is forwarded to the employee's manager who can approve/reject the application (Fig. \ref{fig:BPM_UIs}(b)) or request additional information/changes to the application (Fig. \ref{fig:BPM_UIs}(c)). If approved by the manager, the application is forwarded to the director who makes the final decision on the travel request.

\subsection{Agents in \textit{Travelbot}}
\textit{Travelbot} contains multiple agents that enable it to respond to a broad range of user requests. 
The 3S orchestrator coordinates between the composed agents to respond to the user or execute any actions requested by the user. Table \ref{tab:travelbot_convo} includes a sample interaction between \textit{Travelbot} and a manager. 

\subsubsection{Chit-Chat Agent}
A simple dialog agent, Chit-Chat is implemented as a question answering dialog tree that is capable of answering utterances related to greetings (e.g. ``hello'', ``how are you'', ``goodbye'', etc.). It can also introduce the assistant answering questions such as ``who are you'' and ``how can you help me''. Additional intents and dialog trees are also included to make the assistant more human-like and improve the user experience -- e.g. ``what are you doing'', ``tell me a joke'', etc. This agent does not have any act skills; it does not cause any side effects when executed and hence has identical preview and execute modes.

\subsubsection{Publication Query Agent}
This information retrieval agent queries a database containing the publications of employees and conference information. In the adopted use case, this agent can be used to auto-fill the information in the application form related to travel justification and conference location and dates. It can also be used by managers and directors to obtain information necessary for making decisions. For example, travel requests would be approved for employees presenting their work at the conference as first authors. This agent makes this information accessible to managers through a conversational interface and eliminates the need for them to switch to a different interface. The static composition of this agent consists of a Natural Language Understanding (NLU) understand skill, an act skill capable of retrieving employee information from an employee directory that would be used by the publication query skill (second act skill) to retrieve publication/conference information. Finally, the respond skill generates the final response that will be returned to the user. 

\subsubsection{Business Process Data Query Agent}
Another informational retrieval agent, the business process data query agent allows managers and directors to query the existing travel request applications. Managers and directors can ask questions like ``How many pending travel requests are in my queue?'' or ``How many applications have been submitted by employee X?''. Such queries factor into the decision making process and allow decision makers to make more informed decisions. This agent is composed of three skills: an NLU understand skill, a data query act skill, and a respond skill. 

\subsubsection{Business Process Task Execution Agent}
This agent allows users to move the business process forward by requesting the assistant to submit an application on their behalf (e.g. ``Submit an application to AAAI 2020 for me.'') or approve/reject/send back an application (e.g. ``Reject Jack's application to ICML 2019.''). The agent composition is more complex in this case and consists of four act skills that may or may not be invoked depending on the recognized intent. Like the other agents, this agent also has an intent and entity recognition understand skill and a respond skill. Its act skills are employee directory query, paper directory skill, travel pricing skill and business process task execute act skill. The pricing skill is used auto-fill the ``requested amount'' field in Fig. \ref{fig:BPM_UIs}(a).  

\subsubsection{Travel Estimation Agent}
This agent can respond to inquiries about travel such as -- ``What is the cheapest flight from BOS to SFO leaving on 2019/12/01 and returning on 2019/12/07?''. This information retrieval agent consists of an intent and entity recognition understand skill, a travel pricing skill which uses the Amadeus API to estimate the cost of flights and hotels, and a respond skill that formulates the answer returned to the user. Managers or directors can use this agent to verify the ``requested amount'' field submitted by the employee.  

\subsubsection{Visualization Agent}
Beyond natural language responses, the visualization agent provides the assistant with the ability to return visual responses. This is useful in situations where information retrieval agents return data that is easier to digest in a plot form. Therefore, the user can ask the assistant to plot the results. This agent is composed of an intent and entity recognition understand skill to if and which plot the user wants and what variables to use if multiple options exist. A plotter skill converts the plottable data to an image of a bar graph or doughnut chart or other visualization format that is then forwarded to the user using the respond skill. Due to the stateless implementation of the orchestrator, the data is persisted for a finite number of iterations in the context that is passed between agents the orchestrator. If the visualization agent is invoked by a user utterance, it will look for plottable data in the context. If this data does not exist, then a default response of ``There is no data to be plotted.'' is returned to the user.  

\subsubsection{Alerting using Natural Language Query}
\textit{TravelBot} allows the user to create and customize alerts based on user's data-store. The system accepts user input and if it is determined that intent is the creation of alert, the sentence is then passed through an NLQ module \cite{sen2019natural}, \cite{sen2018functional}, \cite{jammi2018tooling}. The NLQ module can express any natural language sentence into a data query which can run on top of many different types of data-stores (ElasticSearch, SQL, etc). The query from NLQ module is then stored and a daemon runs in the background monitoring these queries. If there is any change in the data store (Addition of new row, changes in values of an existing row) and the query output changes, an alert is generated and sent to the user on the communication channel. The system determines the communication channel for a specific user through conversation. Alerts could be configured to send notifications via Slack, email or SMS using the Twilio API. 

\section{Use Case -- 2: Loan Application}
\begin{figure}[h]
    \centering
    \includegraphics[width=\linewidth]{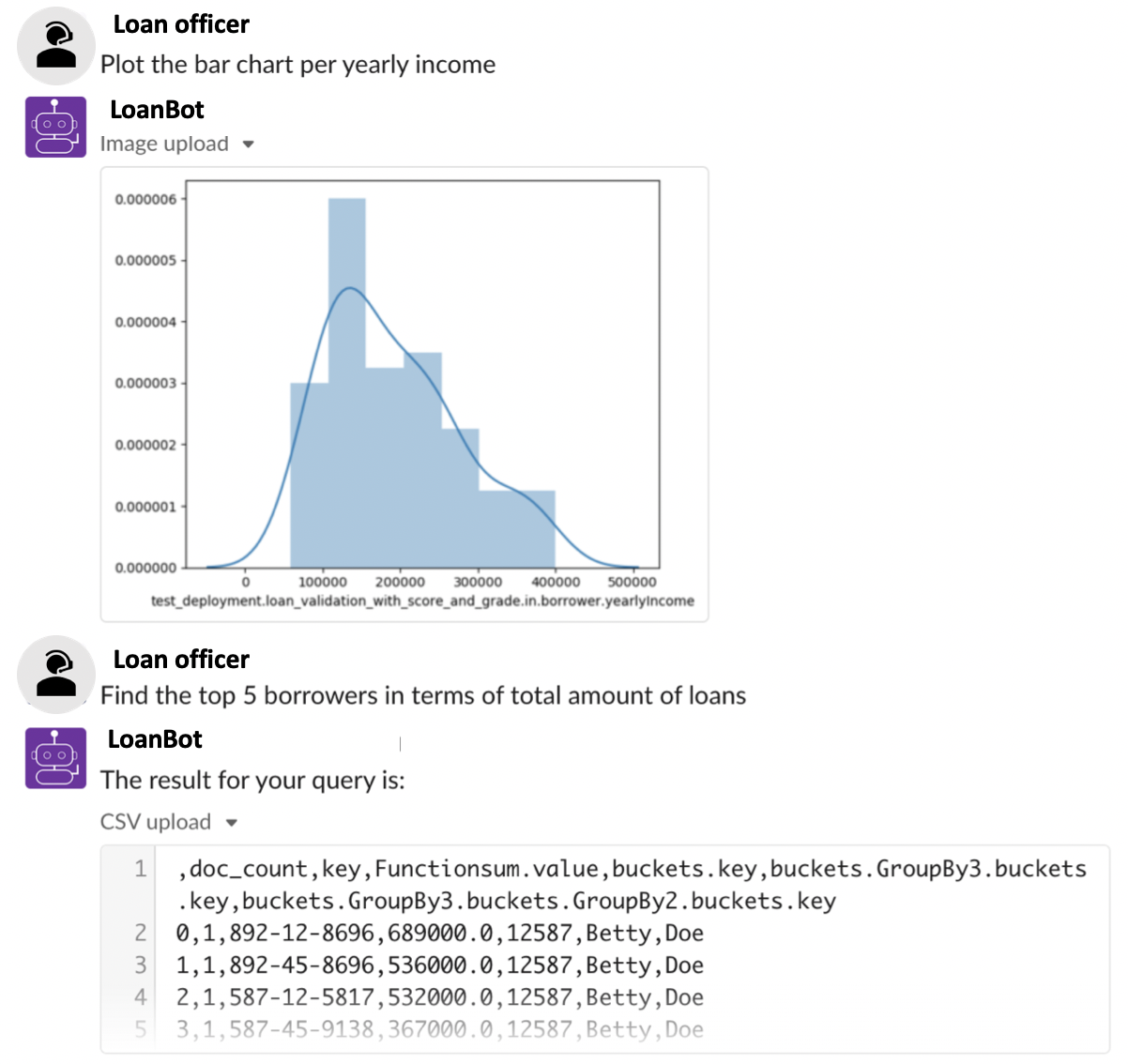}
    \caption{Multi-modal responses from the \textit{LoanBot}.}
    \label{fig:loan_convo}
\end{figure}

Using the same multi-agent framework, we also develop a loan officer chatbot for a loan application business process. We include: 1) a data query agent capable of answering questions related to existing applications, 2) a business rules agent that can provide insights into whether an application will be accepted or rejected, 3) a visualization agent capable of representing data in different formats, 4) an alerts agent that sends notifications when changes to loan applications occur, and 5) a content analyzer that can process loan documents and extract information from them. Sample conversations with this chatbot are included in Table \ref{tab:loanbot_convo} and Fig. \ref{fig:loan_convo}. 

\begin{table}[h]
\centering
\small
    \begin{tabular}{p{6cm} p{2cm}}
         \textbf{Conversation} & \textbf{Responding agent} \\
         \toprule
         Loan Officer: What is the total loan amount for borrowers with credit score more than 500? & \multirow{2}{*}{Data query}\\
         \textit{Loanbot}: The sum value is 137368000.0 & \\
         \midrule
         Loan Officer: Who are the top 3 borrowers with average amount more than 10000 & \multirow{2}{*}{Data query}\\
         \textit{Loanbot}: These are the value: 1). average of J. Smith is 584917\$, 2). average of V. Doe is 575692\$, 3). average of Y. Doe is 557615\$ & \\
         \midrule
         Loan Officer: List all borrowers with yearly income more than 50000 but credit score less than 150 & \multirow{2}{*}{Data query}\\
         \textit{Loanbot}: Total records found are 82. Here is the link: $<$url$>$ & \\
         \midrule
         Loan Officer: Plot the bar chart per yearly income
 & \multirow{2}{*}{Visualization}\\
         \textit{Loanbot}: $<$image$>$ (Fig. \ref{fig:loan_convo}) & \\
         \midrule
         Loan Officer: Find the top 5 borrowers in terms of total amount of loans & \multirow{2}{*}{Data query}\\
         \textit{Loanbot}: The result for your query is: $<$csv$>$ (Fig. \ref{fig:loan_convo}) & \\
         \midrule
         Loan Officer: Could you process an application requesting a loan of 3000\$? & \multirow{2}{*}{Business rules}\\
         \textit{Loanbot}: What is the credit score? & {} \\
         \midrule
         Loan Officer: 400  & \multirow{2}{*}{Business rules}\\
         \textit{Loanbot}: What is the annual salary (in USD) & {} \\
         \midrule
         Loan Officer: 5000  & \multirow{2}{*}{Business rules} \\
         \textit{Loanbot}: In how many months will the loan be paid back? &  \\
         \midrule
         Loan Officer: 12  &  \multirow{2}{*}{Business rules}\\
         \textit{Loanbot}: High risk loan. This loan request should not be approved &  \\
         \bottomrule
    \end{tabular}
    \caption{Sample conversation with LoanBot}
    \label{tab:loanbot_convo}
\end{table}

% \begin{figure}
%     \centering
%     \includegraphics[width=0.8\linewidth]{business_rules.png}
%     \caption{Conversation with LoanBot: business rules agent.}
%     \label{fig:business_rules}
% \end{figure}

% The OCR agent extracts text from images that are submitted by the user. In the travel use case, this could be extracting the expiry date of a passport, passport number for flight reservation, etc. The agent is composed of three skills: an intent and entity recognition, an OCR skill (can be any state-of-the-art algorithm) and a respond skill that confirms the values with the user. This agent allows the assistant to handle inputs other than text, images in this case. 
% \subsection{BAI NLQ Agent}
% The Business Automation Insights (BAI) data lake can be queried using an NLQ framework \cite{sen2019natural}, \cite{sen2018functional}, \cite{jammi2018tooling}. This information retrieval agent does not require an intent recognition understand skill since the scope of intents it can cover is too broad. Instead, the utterance is directly forwarded to the NLQ skill that serves as both the understand and act skill. This NLQ skill converts the natural language statement to an ElasticSearch query and then runs this query to obtain the results in the form of a json file. Finally, a json file parser extracts the sought-after information and a respond skill wraps the information in a natural language sentence that is returned to the user. 

\section{Challenges}
Multiple challenges have hindered the deployment of autonomous conversational agents in BPA domains. We focus on a few of these challenges and discuss how the proposed framework can overcome them. 

\subsubsection{Coding Overhead}
The enterprises that need chatbots often do not have the technical expertise or resources to maintain and improve the framework. Thus, it is important that the system requires minimal coding overhead to deploy in new domains, expand the scope of the system and improve its performance by adding new agents during its lifetime. Our framework assumes that independent developers, whether domain experts or software developers, have created the autonomous agents but require that they abide by the adopted input-output contracts. This reduces the coding overhead to integrate the agents. Furthermore, the 3S orchestrator is robust enough to handle the noise in the agent confidence which is a byproduct of independent agent developers.

\subsubsection{Scalability}
Expanding the scope and capabilities of a conversational agent requires significant developer effort and may result in performance degradation. In the proposed framework, integration is trivial but naively adding agents may result in performance issues if the orchestrator incorrectly selects agents. Adopting an orchestration model that learns from previous interactions and changes in the system would improve the overall performance of the assistant. 

\subsubsection{Agent Overlap}
As the number of agents in the framework increases, some agent functionality and knowledge may overlap. This may cause conflicts among agents when attempting to respond to users and degrade overall performance. As a result, it is important to model or quantify agent overlap and incorporate it within the orchestrator model to insure that the orchestrator will select the best agent to complete the task of responding to the user. Adopting a learning selector in the orchestrator pipeline can lead to better selection models that can handle agent overlap and prevent performance degradation. 

\subsubsection{Multi-user Support}
Giving users the ability to compose their personalized assistants requires the framework to support multiple users. While this can be resolved by creating a session per user, it provides an opportunity to increase the available data from which the orchestrator can learn better execution models. Therefore, the proposed framework should consider approaches to learn from multiple users' data while maintaining privacy and security of the data.  

\subsubsection{Access Control}
With different personas in business processes, some agents may not be accessible to specific users. For example, with the \textit{Travelbot}, employees should not have access to the "approve/reject" functionality. Our framework requires that these restrictions are implemented inside the agent. When the user submits a request, the orchestrator forwards this request to all agents and expects them to decline execution if the user does not have the required access.

\section{Conclusion}
In this paper, we presented a multi-agent framework to develop a conversational assistant supporting multiple capabilities such as querying data through natural language, autonomously executing tasks in a business process, alerting users of changes in the business process and visualizing data in various forms based on the user's requests. An orchestrator capable of combining independently developed skills with minimal overhead is the key to effectively deploy more powerful conversational assistants. While many challenges remain such as scalability and agent overlap, preliminary results motivate follow on research.
%Future directions include creating a stateful orchestrator capable of supporting more complex conversational constructs (such as co-reference resolution and disambiguation) and agent cooperation. Incorporating user feedback to learn from mistakes can also be incorporated into the orchestration model. Enabling the orchestrator to compose agents to respond complex user inputs should also be investigated. 

\bibliographystyle{aaai}
\bibliography{BPA_chatbot}

\end{document}